\documentclass[10pt,twocolumn,letterpaper]{article}

\usepackage{iccv}
\usepackage{times}
\usepackage{epsfig}
\usepackage{graphicx}
\usepackage{tabularx}
\usepackage{algorithm}
\usepackage{algorithmic}
\usepackage{amsmath,mathtools}
\usepackage{amssymb}
\usepackage[inline]{enumitem}
\usepackage[dvipsnames,table]{xcolor}
\usepackage{booktabs}
\usepackage{tabularx}
\usepackage{multirow}
\usepackage{colortbl}
\usepackage{subcaption}
\usepackage{bbm}
\usepackage{caption}

\usepackage{diagbox}
\makeatletter
\@namedef{ver@everyshi.sty}{}
\makeatother
\usepackage{tikz}

\usepackage{algorithm}
\usepackage{listings}

\usepackage{etoolbox}
\makeatletter
\AfterEndEnvironment{algorithm}{\let\@algcomment\relax}
\AtEndEnvironment{algorithm}{\kern2pt\hrule\relax\vskip3pt\@algcomment}
\let\@algcomment\relax
\newcommand\algcomment[1]{\def\@algcomment{\footnotesize#1}}
\renewcommand\fs@ruled{\def\@fs@cfont{\bfseries}\let\@fs@capt\floatc@ruled
  \def\@fs@pre{\hrule height.8pt depth0pt \kern2pt}%
  \def\@fs@post{}%
  \def\@fs@mid{\kern2pt\hrule\kern2pt}%
  \let\@fs@iftopcapt\iftrue}
\makeatother
\usepackage{xspace}
\usepackage{bm,upgreek}
\usepackage[pagebackref=true,breaklinks=true,letterpaper=true,colorlinks,citecolor=ForestGreen, bookmarks=false]{hyperref}

\def\HiLi{\leavevmode\rlap{\hbox to 0.85\linewidth{\color{gray!20}\leaders\hrule height .8\baselineskip depth .5ex\hfill}}}

\makeatletter
\newcommand{\algcolor}[2]{%
  \hskip-\ALG@thistlm\colorbox{#1}{\parbox{\dimexpr\linewidth-2\fboxsep}{\hskip\ALG@thistlm\relax #2}}%
}

\makeatother

\makeatletter

\newcommand*\wthelper[2]{%
        \hbox{\dimen@\accentfontxheight#1%
                \accentfontxheight#11.3\dimen@
                $\m@th#1\widetilde{#2}$%
                \accentfontxheight#1\dimen@
        }%
}

\newcommand*\accentfontxheight[1]{%
        \fontdimen5\ifx#1\displaystyle
                \textfont
        \else\ifx#1\textstyle
                \textfont
        \else\ifx#1\scriptstyle
                \scriptfont
        \else
                \scriptscriptfont
        \fi\fi\fi3
}
\makeatother

\newcommand{\ra}[1]{\renewcommand{\arraystretch}{#1}}

\definecolor{Gray}{gray}{0.85}
\newcolumntype{g}{>{\columncolor{Gray}} c}

\newcommand{\vv}{{\bm v}}

\newcommand{\yy}{{\bm y}}

\newcommand{\zz}{{\bm z}}

\newcommand{\barww}{{\bm {\hat w}}}
\newcommand{\barzz}{{\bm {\hat z}}}

\graphicspath{{./figures/}}
\definecolor{gtable}{rgb}{0.0, 0.5, 0.0}

\newcommand{\system}{THAT\xspace}
\newcommand{\xadv}{X_{adv}}
\newcommand{\uu}{{\bm u}}

\newcommand*{\aka}{\emph{a.k.a}\@\xspace}

\iccvfinalcopy % *** Uncomment this line for the final submission

\def\iccvPaperID{****} % *** Enter the ICCV Paper ID here

\ificcvfinal\fi

\begin{document}

%%%%%%%%% TITLE
\title{THAT: \underline{T}wo \underline{H}ead \underline{A}dversarial \underline{T}raining for Improving Robustness at Scale}

\author{Zuxuan Wu$^{1,3}$ \qquad Tom Goldstein$^{2}$ \qquad Larry S. Davis$^{2}$ \qquad Ser-Nam Lim$^{3}$ \\
$^{1}$~Fudan University \qquad $^{2}$~University of Maryland  \qquad $^{3}$~Facebook AI 
}

\maketitle

%%%%%%%%% ABSTRACT
\begin{abstract}
Many variants of adversarial training have been proposed, with most research focusing on problems with relatively few classes.   In this paper, we propose Two Head Adversarial Training (THAT), a two-stream adversarial learning network that is designed to handle the large-scale many-class ImageNet dataset. The proposed method trains a network with two heads and two loss functions; one to minimize feature-space domain shift between natural and adversarial images, and one to promote high classification accuracy.  This combination delivers a hardened network that achieves state of the art robust accuracy while maintaining high natural accuracy on ImageNet. Through extensive experiments, we demonstrate that the proposed framework outperforms alternative methods under both standard and ``free'' adversarial training settings.   
\end{abstract}

\section{Introduction}
Convolutional neural networks have demonstrated remarkable performance in a multitude of computer vision tasks like image classification~\cite{he2016deep, xie2017aggregated}, object detection~\cite{ren2015faster,carion2020end, focalloss}, \etc. 
Despite the power of these state-of-the-art models, they are found to be extremely unstable to input perturbations~\cite{hendrycks2019benchmarking,Kurakin2016,geirhos2018generalisation}. This fragility can be exploited by crafting adversarial examples, which are optimized to manipulate networks while appearing innocuous to humans. 

One popular method to mitigate the brittleness of neural networks is adversarial training~\cite{madry2018towards,Goodfellow2014,kannan2018adversarial}, in which network parameters are updated using adversarially perturbed images. This produces a hardened network that is robust to adversarial perturbations in the pixel space. While adversarial training is able to increase the robustness of classifiers, it often reduces accuracy on clean images at test time~\cite{zhang2019theoretically,tsipras2018robustness,balaji2019instance}. This accuracy loss is believed to be in part because of fundamental tradeoffs between accuracy and robustness \cite{shafahi2018adversarial,tsipras2018robustness,balaji2019instance}, and in part because of domain shift between clean and adversarial image distributions~\cite{xie2020adversarial,mao2019metric, dubey2019defense}. 

\begin{figure}[t]
   \centering
      \includegraphics[width=1.0\linewidth]{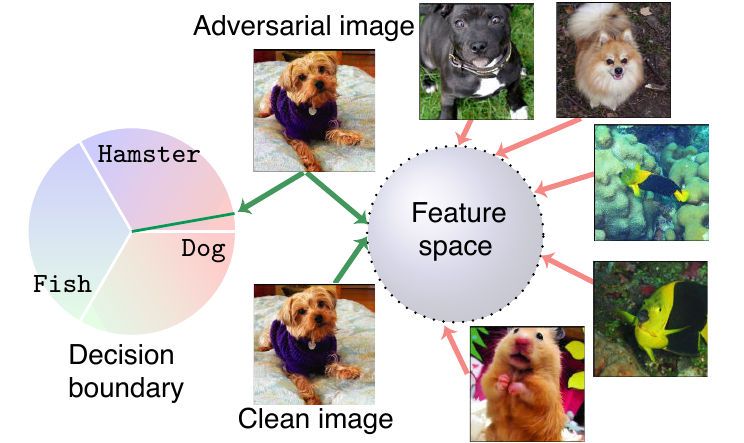}
      \vspace{-0.1in}
           \caption{\textbf{A conceptual overview of our framework.} We train a neural network with two heads, each with its own loss function.  The first loss term acts on the ``feature extraction head,''  and promotes high feature similarity between an adversarial image and its clean counterpart.  The second loss term acts on the ``classification head,'' and enforces that adversarial images receive  correct class labels.}
\end{figure}

In light of this, we propose to align features of clean images with their adversarially perturbed counterparts for improved clean accuracy and robustness.  Our approach is inspired by recent self-supervised learning frameworks~\cite{chen2020simple,he2020momentum}, in which two random crops of the same image are considered as a positive pair and their distance in feature space is minimized to learn generic feature representations. Here, a clean image and its adversarial copy form a positive pair, and by contrasting it with other pairs, we modulate standard adversarial training to prevent the adversarial feature distribution from drifting away from the natural distribution.

With this in mind, we introduce Two Head Adversarial Training (THAT), an adversarial training framework that uses multiple training objectives to boost robustness. \system adversarially trains a robust encoder using two parallel heads, a feature head and a classification head, both sitting on a shared ResNet backbone.  For each training step, clean images are passed through a naturally trained clean network to produce features. Then, adversarial examples are made by attacking the classification head of the robust encoder.  A loss function is then computed that contains two terms: a contrastive loss that promotes similarity between the natural features from the clean encoder and the adversarial features from the robust encoder, and a classification loss from the classification head.  The former term enforces that feature representations of adversarial images are aligned with natural feature representations (\ie, domain shift is minimized), while the second term ensures that the feature representations contain the information needed for accurate classification.  The robust encoder is then updated to minimize the combined loss.

At test time, we consider two different classification modes for defense.  In addition to using a standard classification head, we also consider a nearest-neighbor-based classification procedure that relies on the feature extraction head.

We conduct extensive experiments on ImageNet~\cite{imagenet_cvpr09} using standard adversarial training updates, and also using accelerated (\aka, ``free'' \cite{shafahi2019adversarial}) adversarial updates. In both settings, we demonstrate that \system outperforms other adversarial training frameworks in terms of both clean accuracy and robustness with different backbones.

\section{Background and Related Work}
\noindent \textbf{Adversarial robustness.} To mitigate threats posed by adversarial examples~\cite{Moosavidezfooli2016,Szegedy2013}, adversarial training~\cite{Goodfellow2014,madry2018towards,shafahi2019adversarial,wong2020fast,engstrom2018evaluating,zhang2019defense,Xie2020intriguing} solves a min-max optimization problem in which adversarial examples are crafted to maximize the training loss, and these examples are then used to update network parameters during loss minimization.   This process can be interpreted as approximately solving the saddle-point optimization problem:
\begin{align} 
   \min_{\bm\theta}\mathbb{E}_{(X,y)\sim\mathcal{D}}\left[ \max_{\|\mathbf{\delta}\|_{\infty}<\epsilon}\ell(f(X+\mathbf{\delta}),\,y;\,{\bm \theta}) \right],
\end{align}
where $X$ is a clean image drawn from the training set $\mathcal{D}$ with $y$ as its label, $f$ represents a model parameterized by weights $\bm \theta$, $\delta$ is the adversarial perturbation, and $\ell$ denotes the cross-entropy loss.% The problem is solved by an alternative stochastic approach---iteratively samples a mini-batch of training data, perturbs images in the batch adversarially with $K$ steps PGD, and then takes minimization steps to update network parameters.

Early adversarial training methods crafted adversarial examples with the Fast Gradient Sign method (FGSM)~\cite{Goodfellow2014}, but this approach results in networks that are easily broken by multi-step attacks.
 Madry \etal use project gradient descent (PGD) to generate adversarial examples to improve robustness for multi-step attacks~\cite{madry2018towards}. PGD performs gradient ascent on input images in the signed gradient direction with respect to the classification loss, and then clips the perturbation to enforce an $\ell_{\infty}$-norm constraint.
Kannan \etal further regularize adversarial training by penalizing the difference between logits from clean images and their adversarial variants~\cite{kannan2018adversarial}. Xie \etal introduce denoising blocks that use self-attention in feature maps to improve adversarial training~\cite{xie2018feature}. Balaji \etal improve adversarial training for ImageNet by imposing different robustness criteria for each training sample \cite{balaji2019instance}.
 The TRADES method achieves high levels of robustness by training with a loss that promotes similarity between the predicted label scores for natural and adversarial images~\cite{zhang2019theoretically}. Due to the huge computational overhead of adversarial training methods, Shafahi \etal introduce a ``free'' adversarial training strategy that updates model weights and generates adversarial examples by replaying mini-batches~\cite{shafahi2019adversarial}.

 In our work, we build upon standard and free adversarial training settings and demonstrate that explicitly contrasting features improves both clean accuracy and robustness. Dubey~\cite{dubey2019defense} \etal show that nearest neighbor search can be used for deflecting strong attacks, but this requires an extremely large gallery of reference images ($\sim 1$ billion) to achieve good performance. Instead, we show that our network is able achieve strong defense via nearest neighbors using the much smaller ImageNet dataset. In addition, Mao \etal use a triplet loss for better adversarial training~\cite{mao2019metric}, which requires manual hard negative mining. Our contrastive loss automatically performs hard negative mining with a number of negative samples.

\vspace{0.05in}
\noindent \textbf{Representation learning with contrastive losses.} Extensive studies have been conducted to learn visual representations in a self-supervised manner~\cite{noroozi2016unsupervised,caron2018deep,larsson2016learning, misra2019self,yan2020cluster,gidaris2020learning,dosovitskiy2016discriminative}. Among these approaches, contrastive learning methods~\cite{wu2018unsupervised,he2020momentum, chen2020simple,tian2019contrastive} currently achieve state-of-the-art results by maximizing the agreement of positive pairs (two random crops of the same image) relative to a large number of negative pairs. Wu \etal perform instance discriminative tasks based on the entire ImageNet training set ~\cite{wu2018unsupervised}. MoCo introduces a memory bank to maintain consistent representations of negative samples with the help of a momentum encoder~\cite{he2020momentum}. In this paper, we focus on improving the robustness of neural networks with contrastive losses. We consider a clean image and its adversarially perturbed variant as a positive pair rather than two stochastic data augmentations.

\vspace{0.05in}
\noindent \textbf{Adversarial training and self-supervised learning.} There are some very recent studies exploring adversarial training with self-supervised learning~\cite{ho2020contrastive,jiang2020robust,kim2020adversarial,DBLP:conf/cvpr/Chen0C0AW20}. Chen \etal use several self-supervised losses like predicting rotations, permutations, to pretrian a ResNet and study its robustness~\cite{DBLP:conf/cvpr/Chen0C0AW20}. \cite{ho2020contrastive,jiang2020robust,kim2020adversarial,DBLP:conf/cvpr/Chen0C0AW20} focus on better pretraining with adversarial examples for downstream tasks on small datasets. In contrast, we use a contrastive branch to align features of clean and adversarial images to boost the performance of standard adversarial training. In addition, we also target at large datasets for large-scale adversarial training. 

%%%%%%%%% BODY TEXT
\begin{figure*}[t]
   \centering
      \includegraphics[width=0.85\linewidth]{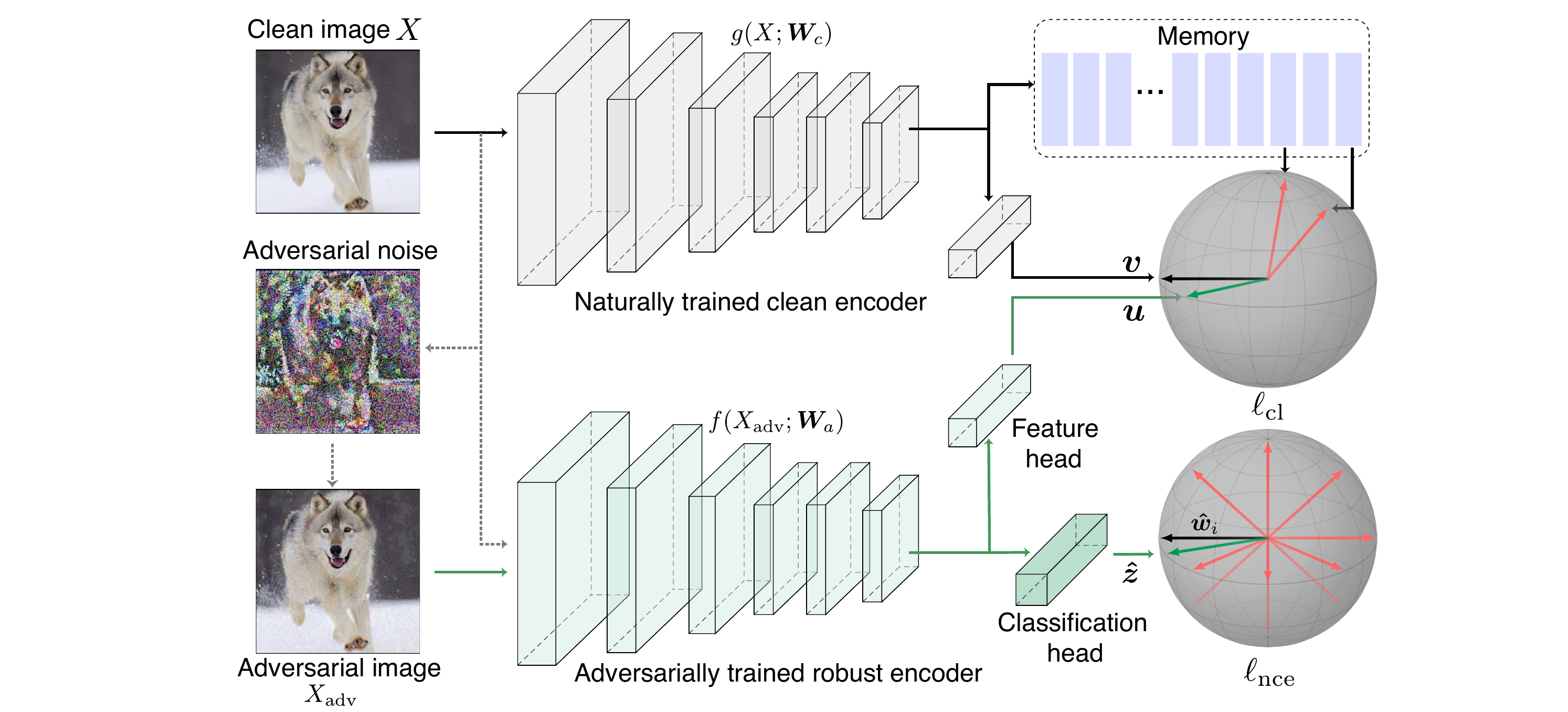}
      \vspace{-0.1in}
           \caption{\textbf{An overview of the framework.} Given a pair containing a clean image and its adversarial variant, features from both images are mapped to locations on a hypersphere.  After robust optimization, both feature vectors lie in close proximity. At the same time, the classification logits from the clean and adversarial image are projected onto a sphere, and forced to lie close to the ground truth label during training. See texts for more details.}
           \label{fig:framework}
\end{figure*}

\vspace{0.05in}
\noindent \textbf{Theoretical motivation and relation to TRADES.}\label{sec:trades}
A successful and well-known training strategy for adversarial defense is TRADES \cite{zhang2019theoretically}.   
This method is related to \system in that it can be interpreted as applying a regularization to compare logits (as opposed to features).  However, while TRADES has been highly successful at robust optimization for CIFAR-10 and MNIST, which have relatively few classes, we have found the TRADES objective to lead to unstable and non-convergent behavior when used on ImageNet.  This problem was also observed in \cite{qin2019adversarial}.  Here, we explore theoretical reasons for the failure of TRADES and discuss how these weaknesses motivate the use of \system. 

TRADES is based on the decomposition $\mathcal{R}_{rob} = \mathcal{R}_{nat} + \mathcal{R}_{bnd},$ which represents the robust accuracy of a model as the sum of the natural risk (\ie, the 0-1 loss), and the ``boundary'' risk.  The boundary risk is the probability that, for an input $X$, there exists an input point $X'$ within a ball around $X$ such that $X$ and $X'$ are assigned different labels.  These terms are then upper bounded by a smooth loss and minimized.  The multi-class TRADES objective is:
\begin{align} \label{trades}
   \min_{\bm\theta} \quad & \mathbb{E}_{(X,y)\sim\mathcal{D}} \,\,
   				 \,\ell(f(X),\,y;\,{\bm \theta})\\
				 &+
 \lambda \max_{\|\mathbf{\delta}\|_{\infty}<\epsilon} \mathcal{L}(f(X+\mathbf{\delta}),f(X);\,{\bm \theta}),
\end{align}
 where $\mathcal{L}$ is a loss function that measures the discrepancy between class label distributions predicted for $X$ and its adversarial example $X+\mathbf{\delta}$, and $\lambda$ is a scalar parameter.
 
For 2-class problems, the loss \eqref{trades} is a tight convex relaxation of the non-convex 0-1 loss whenever $\mathcal{L}$ is a calibrated loss function that satisfies several weak assumptions \cite{zhang2019theoretically}.  However, the tightness result for TRADES is specific to binary classification; the relaxation is no longer tight for multi-class problems, and becomes extremely pathological for many-class problems like ImageNet. 
 
To understand why this is, observe that TRADES uses the KL divergence for $\mathcal{L}(\cdot,\cdot)$ when solving multi-class problems. If $p$ and $q$ are the class distributions predicted by the network for $X$ and its adversarial example, respectively, then the TRADES loss contains the KL-divergence term:
    \begin{equation} \label{kl}
    \mathcal{L}(f(X+\mathbf{\delta}),f(X);\,{\bm \theta}) = \sum_i  p_{i} \log(p_i/q_i),
    \end{equation}
where the sum in Eqn.~\eqref{kl} is over class indices.  For typical ImageNet images,
     the distributions $p$ and $q$ contain just a few relevant/large class probabilities, and then a ``long tail'' with nearly 1000 small probabilities that are generally discarded when making classification decisions.  The KL loss used by TRADES  sums over all of these class labels, and its value becomes highly dominated by the many contributions from classes with small probabilities. Furthermore, the smallest values of $q_i,$ which appear in the denominator of Eqn.~\eqref{kl}, make the loss most sensitive to unlikely classes.  In practice, we often ignore such class labels with near-zero probability, and yet they account for a dominant share of the TRADES objective. 
     
     Our proposed method is similar in spirit to TRADES ---it enforces similarity between network outputs for natural and adversarial images.  But unlike TRADES, our proposed method uses a contrastive loss that is computed using the inner product between feature vectors, and is dominated by large entries in feature representations.  This enables THAT to do contrastive learning while avoiding the pathological behaviors that TRADES suffers in the many-class regime. 
 
\section{\system: Two Head Adversarial Training}
\label{sec:advcon}
In this section, we introduce \system, a contrastive framework with a naturally trained clean encoder and an adversarially trained robust encoder. The method learns feature representations that are robust to attacks while simultaneously achieving high classification accuracy.  Each of these objectives is achieved with a loss function on a separate head of the network.  We describe each component of our framework here.

\vspace{0.05in}
\noindent \textbf{Naturally trained clean encoder.} The natural/clean encoder $g(X; {\bm W_c})$, parameterized by weights ${\bm W_c}$, takes a clean image $X$ as inputs. Following~\cite{chen2020improved}, it uses a two-layer projection head on top of feature maps from the $\texttt{Res\_5}$ stage of a ResNet to produce a 128D feature representation ${\bm v}$. The feature vector is further normalized with an $\ell_2$ norm. We set the weights of the clean encoder from a pre-trained self-supervised model~\cite{chen2020improved}.  The weights of this model remain frozen while the robust model is trained. % so as to guide adversarial training---preventing features of adversarial examples from drifting away from their images. Thus, we fix the weight matrices ${\bm W_c}$ during training, but we also discuss when the weights are updated in the experiments. \tom{So the whole clean encoder pipeline is not trained at all?  Right? \zx{yes, they are fixed}}

\vspace{0.05in}
\noindent \textbf{Adversarially trained robust encoder.} The robust encoder $f(X_{\text{adv}}; {\bm W_a})$,  parameterized by weights ${\bm W_{a}}$, is a two-head architecture with a feature head and a classifier head. In particular, ${\bm W_{a}} = \{{{\bm W_{ab}}, \bm W_{af}}, {\bm W_{ac}}\}$, where $\bm W_{ab}$, $\bm W_{af}$ and $\bm W_{ac}$ represent the weight matrices for the base ResNet model, a feature head with two fully-connected layers, and the classifier head with one fully-connected layer, respectively. Given a clean image $X$, the robust encoder first generates its adversarially perturbed version $\xadv = X + \mathbf{\delta}$ on-the-fly by attacking the classification head using multiple steps of PGD. The feature head then computes a 128 dimensional feature vector $\uu$ for the perturbed image, which is normalized to have unit $\ell_2$ norm.  At the same time, the separate classifier head produces a logit vector $\zz$ for use in the classification loss.

\vspace{0.05in}
\noindent \textbf{The contrastive loss.} The feature extraction head provides adversarial features $\uu$ for the attack image, and the clean network provides a clean image representation $\vv$.  These two representations form a ``positive pair'' and should be aligned by the contrastive loss.  We also form ``negative pairs'' by comparing $\uu$ with representations from randomly selected clean images, which could be sampled from the same mini-batch or from an external memory bank. Here, we use a memory bank as it is has been demonstrated that using a large number of negative pairs is beneficial~\cite{he2020momentum,wu2018unsupervised}.
Finally, we form the contrastive loss function~\cite{sohn2016improved,oord2018representation}: 
\begin{align}
   \small
\ell_{\text{cl}} (\uu) =-\log \frac{\exp (\bm u^T \bm v / \tau)}{\exp (\bm u^T \bm v / \tau) + \sum_{\bm v\in \mathcal{V}_-}\exp(\bm u^T \bm v/\tau)},
\label{eqn:npair}
\end{align}
where $\mathcal{V}_-$ is the set of features from random negative samples.  The parameter $\tau$ is the ``temperature,'' controlling the sharpness of the distribution. The contrastive loss forces the adversarial representation $\uu$ to be closer to its own clean base image than other images.  This suppresses domain shift between the clean and adversarial images in feature space.

\vspace{0.05in}
\noindent \textbf{Classifier loss.} The contrastive learning loss forces feature representations to be invariant to attacks, but it does not measure classification performance of these representations. To get good classification performance, the classifier head of the robust network produces its own training loss.  For the classifier head, we use a ``normalized'' cross entropy~\cite{wang2018cosface,qi2018low,wu2018unsupervised}, which measures the disparity between output logits and ones-hot vectors using a contrastive loss.  This keeps gradient scaling and training dynamics of the classifier loss similar to the contrastive loss on the features.  The normalized cross entropy loss is:

\begin{align}
\ell_{\text{nce}}(\zz) = -y_i \log \frac{\exp (\barzz^T \barww_i / \eta)}{\sum_{i=1}^{C} \exp(\barzz^T \barww_i / \eta)},
\label{eqn:nce}
\end{align}
where $\barzz$ is the normalized logits based on $\zz$, $\barww_i$ is the $i$ the column in $\bm W_{ac}$ representing the normalized weights for the $i$-th class, $y_i \in \mathbb{R}^{\{0, 1\}}$ is the label for the $i$-th class, and $C$ is the total number of classes in the dataset. 
$\eta$ is a learnable parameter to control the sharpness of the distribution. Here, $\barww_i$ is considered as the class prototype, and we are forcing the logit vectors to lie close to it. This is similar in spirit to Eqn~\eqref{eqn:npair}, in which features from an adversarial example are mapped to be close to features of its clean twin.

Finally, the combined training objective of \system can be written as: %\tom{What is $f$?} \tom{Why did you not put an $f$ inside the CL term? The definition of CL is only on features, so you need a feature extractor here?  The contrastive loss require two different feature extractor, right?  Why isn't that shown? \zx{$f,g$ are defined above in L244, L258. I'm not sure using a single $f$ to denote the outputs of two heads is good. Should we put the cl loss inside the expectation. Not sure how to write it properly.}} 
\begin{align} 
   \min_{\bm W_a} \mathbb{E}_{(X,y)\sim\mathcal{D}} & \,\ell_{\text{nce}}(f(X+\mathbf{\delta}),\,y;\,{\bm W_a})    + \nonumber \\ 
   &\ell_{\text{cl}}(g(X), f(X+\mathbf{\delta})  ;{\bm W_a})   \label{eqn:final_obj} \\
  \text{where}\,\,\, \delta =  &\max_{\|\mathbf{\delta}\|_{\infty}<\epsilon} \ell_{\text{nce}}(f(X+\mathbf{\delta}),\,y;\,{\bm W_a}) .\nonumber 
\end{align}

%##################################################################################################
\begin{algorithm}[t]
   \caption{Pseudocode of our approach in PyTorch style.}
   \label{alg:code}
   \algcomment{\fontsize{7.2pt}{0em}\selectfont
   \vspace{-2.em}
   }
   \definecolor{codeblue}{rgb}{0.580,0.337,0.447}
   \lstset{
     backgroundcolor=\color{white},
     basicstyle=\fontsize{7.2pt}{7.2pt}\ttfamily\selectfont,
     columns=fullflexible,
     breaklines=true,
     captionpos=b,
     commentstyle=\fontsize{7.2pt}{7.2pt}\color{codeblue},
     keywordstyle=\fontsize{7.2pt}{7.2pt},
   %  frame=tb,
   }

\begin{lstlisting}[language=python, mathescape=true]
# g: naturally trained clean encoder
# f: adversarially trained robust encoder
# eps: adversarial perturbation epsilon
# K: number of steps for PGD
# mem: memory bank with clean image features

for x, y in loader:  #  x: data, y: labels
   # generate adversarial examples with K-step PGD
   x_adv = PGD_attack(f, x, y, K, eps)

   # compute features for clean and adversarial images
   feat_clean = g.forward(x)
   feat_adv, logits_adv = f.forward(x_adv)
   
   # compute contrastive losses
   loss_cl = cl_loss(feat_adv, feat_clean, mem)
   
   # compute classification losses
   loss_cls = nce_loss(logits_adv, y)

   loss = loss_cls + loss_cl
   loss.backward()
   optimizer.step()
\end{lstlisting}
\end{algorithm}

\begin{table*}[t!]
   \centering
   \ra{1.0}
   \setlength{\tabcolsep}{0pt} % let TeX compute the intercolumn space
\begin{tabular*}{\textwidth}{@{\extracolsep{\fill}\quad}lllllll}
      \toprule
        & Clean & PGD-10 & PGD-30 & PGD-200 & PGD-1000 \\ 
      \midrule
      \multicolumn{1}{l}{R50} \\
     Standard AT & 50.81                             & 47.78                             & 39.31                             & 38.09                             & 37.74                             \\
    Ours  &53.29 (\textcolor{gtable}{+2.48}) & 49.62 (\textcolor{gtable}{+1.84}) & 41.01 (\textcolor{gtable}{+1.70}) & 40.22 (\textcolor{gtable}{+2.13}) & 39.59 (\textcolor{gtable}{+1.85}) \\
      \midrule
      \multicolumn{1}{l}{R101} \\

      Standard AT    & 56.21                             & 51.40                             & 42.68                             & 41.08                             & 40.86                             \\
    Ours   &58.17 (\textcolor{gtable}{+1.96}) & 52.70 (\textcolor{gtable}{+1.30}) & 44.02 (\textcolor{gtable}{+1.34}) & 42.79 (\textcolor{gtable}{+1.71}) & 42.34 (\textcolor{gtable}{+1.48}) \\
      \midrule
      \multicolumn{1}{l}{R152} \\

      Standard AT   &57.61                             & 52.13                             & 43.86                             & 42.73                             & 42.14                             \\
    Ours   &60.38 (\textcolor{gtable}{+2.77}) & 54.69 (\textcolor{gtable}{+2.56}) & 45.67 (\textcolor{gtable}{+1.81}) & 44.51 (\textcolor{gtable}{+1.78}) & 44.22 (\textcolor{gtable}{+2.08}) \\
      \bottomrule
      \end{tabular*}
      \vspace{-0.12in}
      \caption{\textbf{Results and comparisons of our method with standard adversarial training} using different backbone networks.}
      \label{tbl:sat}
\end{table*}

\vspace{0.05in}
\noindent \textbf{Defense strategies.}
Once \system is trained, it is able to defend against strong PGD attacks during testing. We experiment with two different classification modes for defense during testing: (i) standard softmax-based defense using outputs from the classification head of the robust encoder; (ii) nearest-neighbor based defense using features from the feature head of the robust encoder. Below we introduce the nearest-neighbor classifier in detail.

The nearest-neighbor classifier computes feature representations for all training samples with the clean encoder and stores them in a memory bank $\bm M_{\text{train}}$. Given a test image $\hat{X}$, we first compute its feature $\uu_t$ through the robust feature head. Then, the embedding is compared with those of all training samples to retrieve the top-$k$ nearest neighbors. $N_k$ similarity scores are computed based on dot products, and a weighted average of neighbor labels is computed.
More precisely, the confidence of $\hat{X}$ belonging to the $c$-th class is defined as:
\begin{align}
P(c | \hat{X}) = \sum_{i=1}^{N_k} \uu_t^T\uu^i \cdot \yy^i(c),  
\end{align}
where $\uu^i$ is the feature for $i$-th neighbor sample, and $\yy^i$ is the corresponding one-hot label.

\vspace{0.05in}\section{Experiments}
\noindent \textbf{Datasets and metics.} We evaluate our framework on the ImageNet classification benchmark~\cite{imagenet_cvpr09}, which has ${\sim}1.28$ million images annotated into 1000 classes.  We consider two adversarial training settings: (1) standard adversarial training ({\scshape Standard AT})~\cite{madry2018towards,kannan2018adversarial} using $K$-step PGD to generate adversarial examples. This method  increases model robustness but is computationally expensive  (\ie, $K$ times slower compared to natural image training). In this setting, for both training and testing, we consider \emph{targeted attacks} by randomly selecting a targeted class uniformly following~\cite{xie2018feature}; (2) The ``free'' adversarial training method~\cite{shafahi2019adversarial} ({\scshape Free AT}), which speeds up standard adversarial training by updating model weights and generating adversarial examples at the same time with mini-batch replay. Following~\cite{shafahi2019adversarial}, we consider \emph{untargeted attacks} for both training and testing. We report top-1 classification accuracy on the 50k ImageNet validation data using both clean images and adversarially perturbed images and with many-step PGD attacks as in~\cite{xie2018feature,kannan2018adversarial,athalye2018obfuscated}.

\vspace{0.05in}
\noindent \textbf{Implementation details.} We adopt Pytorch for implementation. Since adversarial training on ImageNet is expensive, we use distributed training with synchronized SGD. In particular, for standard adversarial training, we set the maximum perturbation for each pixel to $\epsilon=16$, the step size to $\alpha=4$, and the number of attack iterations to $K=10$. We found that this achieves similar results (\ie, robustness and clean accuracy) compared to using $30$ attack iterations with a step size of $1$, but can reduce training time by 3$\times$. We use a batch of size $4096$ on 32 
Tesla V100 32GB GPUs and train for $100$ epochs as in~\cite{xie2018feature}. The initial learning rate is set to $1.6$, and is decayed by a factor of 10 at the $35$, $60$, and $90$ epoch. For fast adversarial training, we set  $\epsilon=4$ and set the number of replays to $4$, following~\cite{shafahi2019adversarial}. We use a batch size of $2048$ during training and train for $90$ epochs (effectively $23$ epochs with replay~\cite{shafahi2019adversarial}). The clean encoder has the same backbone as the robust encoder, but is initialized from pre-trained self-supervised models~\cite{chen2020improved} and fixed during training. We also show the clean encoder can be trained in Section~\ref{sec:dissussion}.
\begin{figure}[b]
   \centering
   \includegraphics[width=0.9\linewidth]{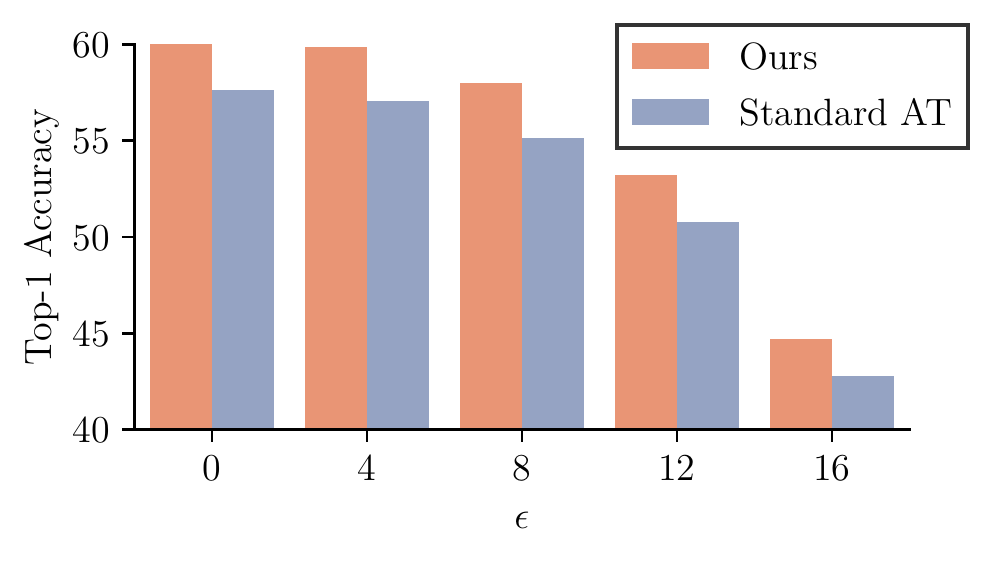}
   \vspace{-0.2in}
   \caption{\textbf{Results of different $\epsilon$} using a ResNet152 architecture, evaluated with PGD-200.}
   \label{fig:eps}
\end{figure}

\subsection{Classifier-based Defense}
\noindent \textbf{Standard adversarial training.} Table~\ref{tbl:sat} summarizes the results of our approach and comparisons with standard adversarial training (standard AT) using three different backbones, \ie ResNet50 (R50), ResNet101 (R101), and ResNet152 (R152). We can see from the table that compared to standard AT, our method with a R152 backbone achieves a clean accuracy of 60.38\% and an accuracy of 44.22\% when evaluated with PGD-1000, offering 2\% (absolute percentage points) compared to the standard AT baseline. We observe similar trends for both R50 and R101, confirming the generalization of our approach with different backbone networks. In addition, the performance of network models degrades when more attack iterations are used and tends to stabilize after 200 iterations. Comparing across different backbones, we can see that models with larger capacity perform better in terms of both clean accuracy and robustness, as observed in~\cite{Xie2020intriguing}. We also evaluate both our approach and standard AT with different maximum perturbation values (\ie $\epsilon$). The results are shown in Figure~\ref{fig:eps}. We see that our approach clearly outperforms the standard baseline model with different perturbation values.

\begin{figure}[t]
   \centering
   \resizebox{1.0\linewidth}{!}{
   \input{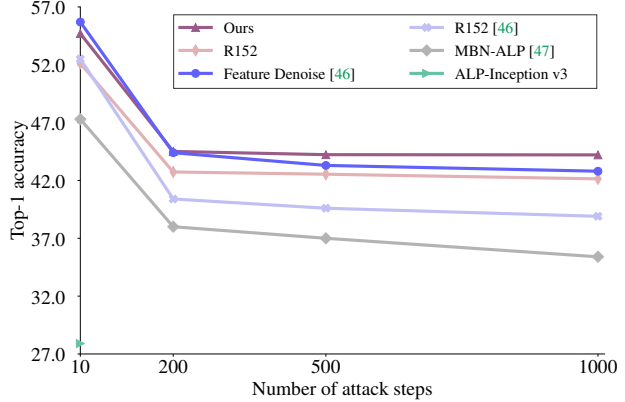}}
   \vspace{-0.3in}
   \caption{\textbf{Comparisons of \system with state-of-the-art models} for standard adversarial training on ImageNet.}
   \label{fig:sota}
\end{figure}

Furthermore, we compare with the following state-of-the-art models on ImageNet: (1) ALP~\cite{kannan2018adversarial}, which penalizes the outputs of clean images to be similar to those of adversarial images with a mean-squared error loss;
(2) Feature Denoising~\cite{xie2018feature}, which adds non-local blocks~\cite{wang2018non} after each residual block in ResNet models
% ~\footnote{Note that we could not replicate the R152 baseline used in~\cite{xie2018feature} in PyTorch, but in our implementation R152 is more robust.}
; (3) MBN-ALP~\cite{Xie2020intriguing}, which reimplements the ALP algorithm by using different batch normalization statistics for clean and adversarial images. Results are summarized in Fig.~\ref{fig:sota}. 

We see from the figure that \system achieves better results than alternative methods when strong attacks are presented at test time (\ie, the number of attack iterations is greater than 200). Compared to Feature Denoising~\cite{xie2018feature} which performs self-attention on feature maps, \system offers 1.4\% gain when evaluated against PGD-1000 and is slightly worse with PGD-10 attacks. We also experiment with non-local blocks and observe that they can slightly improve the performance for weak attacks and clean accuracy. However, adding non-local blocks makes training more computationally expensive as it flattens all the pixels into a huge vector to compute a dense graph. In addition, we also compare our R152 baseline with that in~\cite{xie2018feature}, and our method achieves better robustness against strong attacks while being 0.4\% worse against PGD-10. Furthermore, both \system and Feature Denoising outperform ALP and its variant by clear margins.

\begin{table}[h!]
   \centering
   \ra{1.0}
   % \small
   \setlength{\tabcolsep}{0pt} % let TeX compute the intercolumn space
   \begin{tabular*}{\linewidth}{@{\extracolsep{\fill}\quad}lllll}
   \toprule
     &  Clean & PGD-10 & PGD-100 \\
   \midrule 
   \multicolumn{1}{l}{R50} \\
   Free~\cite{shafahi2019adversarial} &  60.21 &  32.77 &  31.82 \\
   Ours & 63.01 (\textcolor{gtable}{+2.8})  & 34.25 (\textcolor{gtable}{+1.48})& 33.26 (\textcolor{gtable}{+1.44}) \\
   \midrule
   \multicolumn{1}{l}{R101} \\
   Free~\cite{shafahi2019adversarial}  & 63.34 & 35.38 & 34.32  \\
   Ours  & 67.22 (\textcolor{gtable}{+3.88}) & 38.56 (\textcolor{gtable}{+3.18}) & 37.17 (\textcolor{gtable}{+2.85}) \\
   \midrule
   \multicolumn{1}{l}{R152} \\
   Free~\cite{shafahi2019adversarial}  & 64.45 &  36.99 &  35.99 \\
   Ours  & 68.82 (\textcolor{gtable}{+4.37}) & 39.83 (\textcolor{gtable}{+2.84}) & 38.27 (\textcolor{gtable}{+2.28})\\
   \bottomrule
   \end{tabular*}
   \vspace{-0.1in}
   \caption{\textbf{Results and comparisons with the ``free'' adversarial training~\cite{shafahi2019adversarial}}. Here, the PGD attacks are untargeted.}
   \label{tbl:free}
   \end{table}

   \vspace{0.05in}
\noindent \textbf{Accelerated training results.} We now experiment with an accelerated adversarial training scheme to verify that \system is compatible with different strategies. In particular, we train the proposed architecture on top of the ``free'' training framework introduced in~\cite{shafahi2019adversarial}. The results are summarized in Table~\ref{tbl:free}. Similarly to adversarial training, we observe \system offers significant performance gains for both clean accuracy and robustness compared to the baseline method. In particular, R152 offers a 4\% gain for clean accuracy and 2.28\% improvement against PGD-100. Note that the results in Table~\ref{tbl:free} are not directly comparable to Table~\ref{tbl:sat} since the attacks in the ``free'' setting are untargeted. In addition, we only evaluate 100 attack iterations as in~\cite{shafahi2019adversarial} since the performance stabilizes.

\subsection{Nearest neighbor classification}
\label{sec:knn}
We demonstrate that features computed from the feature head of the robust encoder are able to facilitate nearest-neighbor based defense against strong PGD attacks. We compare \system with a 
\textsc{R152-Con} model, which augments a R152 model with a feature head for contrastive learning. R152-Con is trained on clean images without adversarial training, offering an accuracy of 78.89\% on the clean ImageNet validation set.

\begin{table}[h!]
   \centering
   \ra{1.0}
   \setlength{\tabcolsep}{0pt} % let TeX compute the intercolumn space
\begin{tabular*}{\linewidth}{@{\extracolsep{\fill}\quad}*{4}c}
   \toprule
   Method  & PGD-10 & PGD-30 & PGD-500 \\
   \midrule 
   R152-Con  & 15.08 & 11.35 & 10.71 \\
   Ours& 35.38 & 29.24 & 27.67 \\
   \bottomrule
   \end{tabular*}
   \vspace{-0.1in}
   \caption{\textbf{Results and comparisons} of defending with nearest neighbors.}
   \label{tbl:knn}
   \end{table}

Table~\ref{tbl:knn} summarizes the results of \system and R152-Con using the top-50 nearest neighbors to classify test images. 
Interestingly, R152-Con offers a 15\% top-1 accuracy against PGD-10 with nearest neighbor based defense, even though the model is not adversarially trained.  This is much better than the 0.66\%  accuracy of its clean-trained softmax classifier, which indicates that nearest neighbor methods are indeed more resistant to PGD than softmax. 
 \system achieves 35.4\% accuracy against PGD-10, which greatly exceeds naturally trained R152-Con.  To compare to a KNN based robust model,  we report results from~\cite{dubey2019defense} which similarly performs defense with nearest neighbors, but the features are used for classification rather then explicitly designed for contrasting learning. For fair comparisons we follow the setting in~\cite{dubey2019defense}, which constrains perturbations within an $\ell_2$ ball instead of the (more standard) $\ell_\infty$ ball used in other sections of this paper. The results are summarized in Table~\ref{tbl:knn_compare}. Using the same set of images for retrieval (\ie, ImageNet training set with 1.28 million images), \system outperforms \cite{dubey2019defense} by 11.8\% in top-1 accuracy, highlighting the effectiveness of features from the feature head for defense. In addition, with a R152 model, \system achieves performance comparable to Dubey's KNN defense with one billion training images available for retrieval. Note that we did not retrain our network against $L2$ threat models.

   \begin{table}[h!]
      \centering
      \ra{1.0}
      \setlength{\tabcolsep}{0pt} % let TeX compute the intercolumn space
   \begin{tabular*}{\linewidth}{@{\extracolsep{\fill}\quad}lc}
      \toprule
      Method  &  Top-1 accuracy (PGD-10)\\
      \midrule 
      IG-1B-R50~\cite{dubey2019defense} & 46.2 \\
      ImageNet-1.3M (Ours-R152) & 45.6 \\ 
      \midrule
      ImageNet-1.3M-R50~\cite{dubey2019defense} & 23.5  \\
      IMageNet-1.3M (Ours-R50) & 35.3  \\
      \bottomrule
      \end{tabular*}
      \vspace{-0.1in}
      \caption{\textbf{Comparisons with state-of-the-art nearest neighbor defense.} Here, IG-1B denotes the dataset with 1 billion images from Instagram~\cite{mahajan2018exploring}. }
      \label{tbl:knn_compare}
      \end{table}

      \begin{figure}[h]
         \centering
         \includegraphics[width=1.0\linewidth]{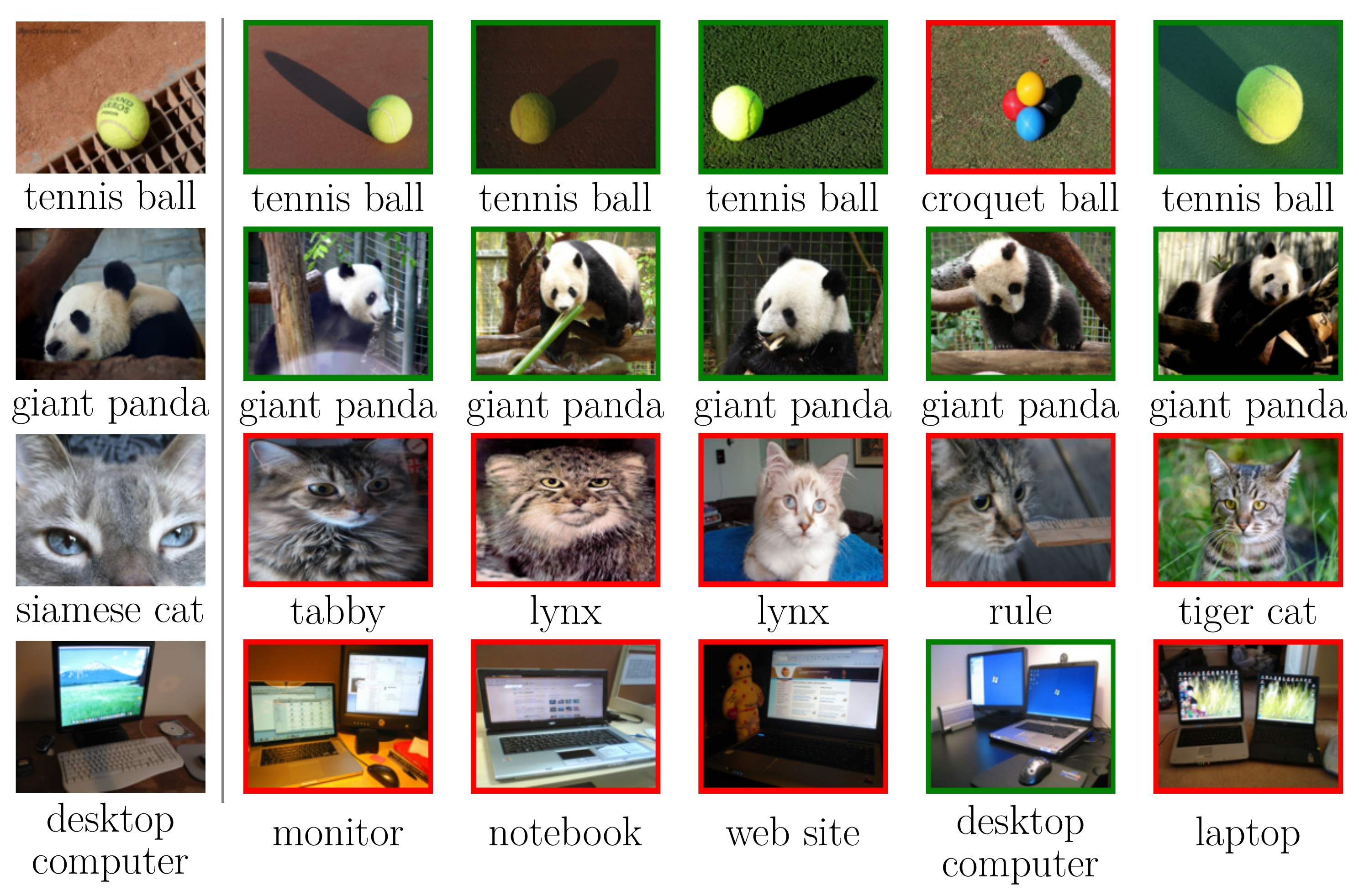}
         \vspace{-0.3in}
                 \caption{\textbf{Top-5 nearest neighbors retrieved given an adversarially perturbed query image.} Left: A query image which has been adversarially perturbed with PGD-10. Right: top-5 nearest neighbors of the query image from the training set of ImageNet.}
                 \label{fig:knn_sample}
      \end{figure}
We also show qualitatively in Figure~\ref{fig:knn_sample} both success (top two rows) and failure (bottom two rows) cases of randomly selected samples with the KNN defense. For both success and failure cases, we see that retrieved samples are indeed visually similar to the query image although the query image is perturbed with adversarial noise. The incorrect predictions for failure cases are largely due to the existence of fine-grained classes, rather than a robustness failure of the embedding. For example, in the third row of Fig.~\ref{fig:knn_sample}, different kinds of cats are retrieved but they belong to different species other than the ``siamese cat''.

\subsection{Discussion}
\label{sec:dissussion}
We conduct a set of experiments to analyze different components of \system, and discuss the results.

\vspace{0.05in}
\noindent \textbf{Ablation on losses of \system.} We show results of \system with different loss functions in Table~\ref{tbl:components}. We observe that compared to our full framework, removing either the contrastive loss or the normalized softmax degrades the performance slightly, yet the resulting models still outperform the standard adversarial training baseline by at least 1\% in clean accuracy and 0.6\% against PGD-1000. This highlights the importance of both components for adversarial training. Furthermore, without the contrastive learning branch, our framework produces 58.92\% and 42.78\% accuracy when evaluated on clean images and against PGD-1000 attacks, respectively. The result is slightly worse compared to removing the normalized softmax (\ie, a standard cross-entropy loss is used in the pipeline), suggesting that contrastive learning is relatively important. 
Note that we did not compare to TRADES~\cite{zhang2019theoretically} because we found it to be unstable and non-convergent, even after hyper-parameter searching. Similar attempts and failures to train TRADES on ImageNet are mentioned in~\cite{qin2019adversarial}.  See Section \ref{sec:trades}.
As a workaround, we compare with standard adversarial training with KL divergence loss to force probability distributions from clean images to be close to those of adversarial images~\footnote{Note that this is different from TRADES as TRADES maximizes the KL divergence to generate adversarial examples, while we instead maximize classification loss as in our approach.}. We see that it achieves high clean accuracy but is extremely vulnerable to strong PGD attacks. This is likely because of the strong emphasis that the KL loss puts on unlikely class labels (see Section \ref{sec:trades}).

\begin{table}[h!]
   \centering
   \ra{1.0}
   \setlength{\tabcolsep}{0pt} % let TeX compute the intercolumn space
   \begin{tabular*}{\linewidth}{@{\extracolsep{\fill}\quad}lccc}
      \toprule
       Method & Clean & PGD-10 & PGD-1000 \\
       \midrule
       Standard AT    & 57.61 & 52.13 & 42.14 \\
       Standard AT  + KL  & 71.05 & 14.13 & 0.00 \\

       \midrule
       Ours w.o. CL & 58.92 & 53.32 & 42.78 \\
       Ours w.o. NCE & 59.64 & 53.80 & 42.77\\
       \midrule
      Ours & 60.38 & 54.69 & 44.22\\
   \bottomrule
   \end{tabular*}
   \vspace{-0.1in}
   \caption{\textbf{Ablating different components of \system}.}
   \label{tbl:components}
   \end{table}

\vspace{0.05in}
\noindent \textbf{Number of negative samples for contrastive learning.} Self-supervised learning methods suggest the number of negative samples used for contrastive learning is important. Therefore, we analyze the performance of \system using different numbers of examples in the memory for contrastive learning. The results are summarized in Table~\ref{tbl:mem}. We see that increasing the number of samples in the memory is indeed beneficial for improving clean accuracy and robustness. With $65536$ samples in the memory, \system offers a clean accuracy of 60.68\%, outperforming standard adversarial training by 3\%.  A memory size of $32768$ offers the best trade-off between clean accuracy and robustness.

\begin{table}[h!]
   \centering
   \ra{1.2}
   \setlength{\tabcolsep}{0pt} % let TeX compute the intercolumn space
   \begin{tabular*}{\linewidth}{@{\extracolsep{\fill}\quad}cccc}
      \toprule
       \# Samples & Clean & PGD-10 & PGD-1000 \\
       \midrule
       0    & 57.61 & 43.86 & 42.14 \\
       \midrule
       4096 & 59.77 & 45.40 & 43.15 \\
       8192 & 58.25 & \textbf{45.89} & 43.79 \\
      16384 & 60.26 & 45.71 & 43.59 \\
      32768 & 60.38 & 45.67 & \textbf{44.22} \\
      65536 & \textbf{60.68} & 45.69 & 43.43 \\
   %   131072 & 59.96 & 45.85 & \textbf{44.24} \\
   \bottomrule
   \end{tabular*}
   \vspace{-0.1in}
   \caption{\textbf{Results of \system} using different number of negative samples.}
   \label{tbl:mem}
   \end{table}

\vspace{0.05in}
\noindent \textbf{Clean encoder.} Instead of freezing the weights of the clean encoder during training, we also experiment with updating its weights to reflect the parameters of the robust encoder with momentum to ensure consistent representations in memory~\cite{he2020momentum}.  The results are presented in Table~\ref{tbl:clean}. This modified implementation that updates both networks at once clearly beats standard adversarial training in terms of robustness, and slightly in terms of clean accuracy. However our proposed framework beats both methods in terms of clean accuracy and robustness. 
   \begin{table}[h!]
      \centering
      \ra{1.0}
      \setlength{\tabcolsep}{0pt} % let TeX compute the intercolumn space
      \begin{tabular*}{\linewidth}{@{\extracolsep{\fill}\quad}lccc}
         \toprule
          Method & Clean & PGD-10 & PGD-1000 \\
          \midrule
          Standard AT    & 57.61 & 52.13 & 42.14 \\
          \midrule
          Ours-MoCo & 57.76 & 53.36 & 43.03 \\
         Ours & 60.38 & 54.69 & 44.22\\
      \bottomrule
      \end{tabular*}
      \vspace{-0.12in}
      \caption{\textbf{Results of updating the clean encoder} in a MoCo~\cite{he2020momentum} fashion.}
      \label{tbl:clean}
      \end{table}

\begin{figure}[h]
   \centering
   \includegraphics[width=0.8\linewidth]{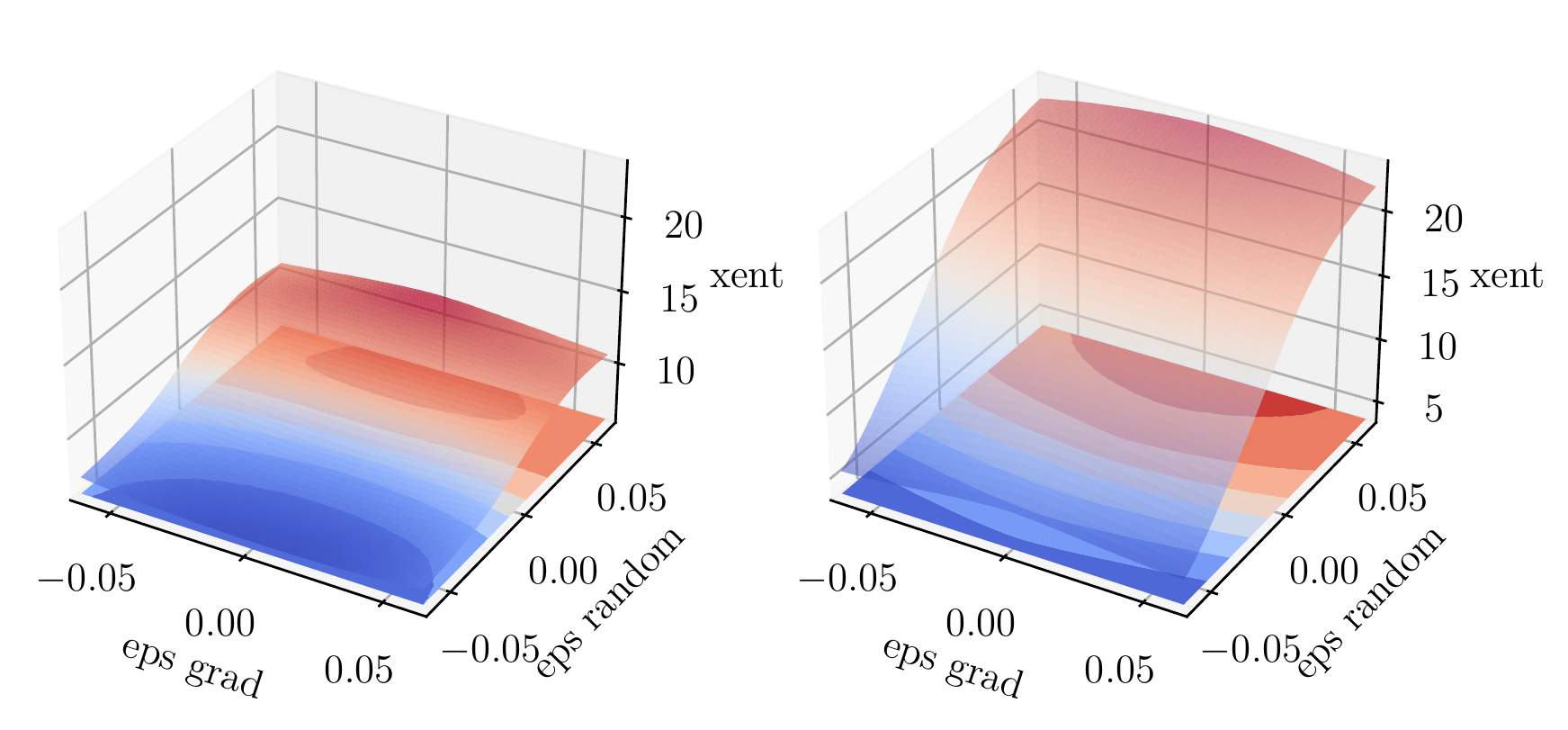}
   \includegraphics[width=0.8\linewidth]{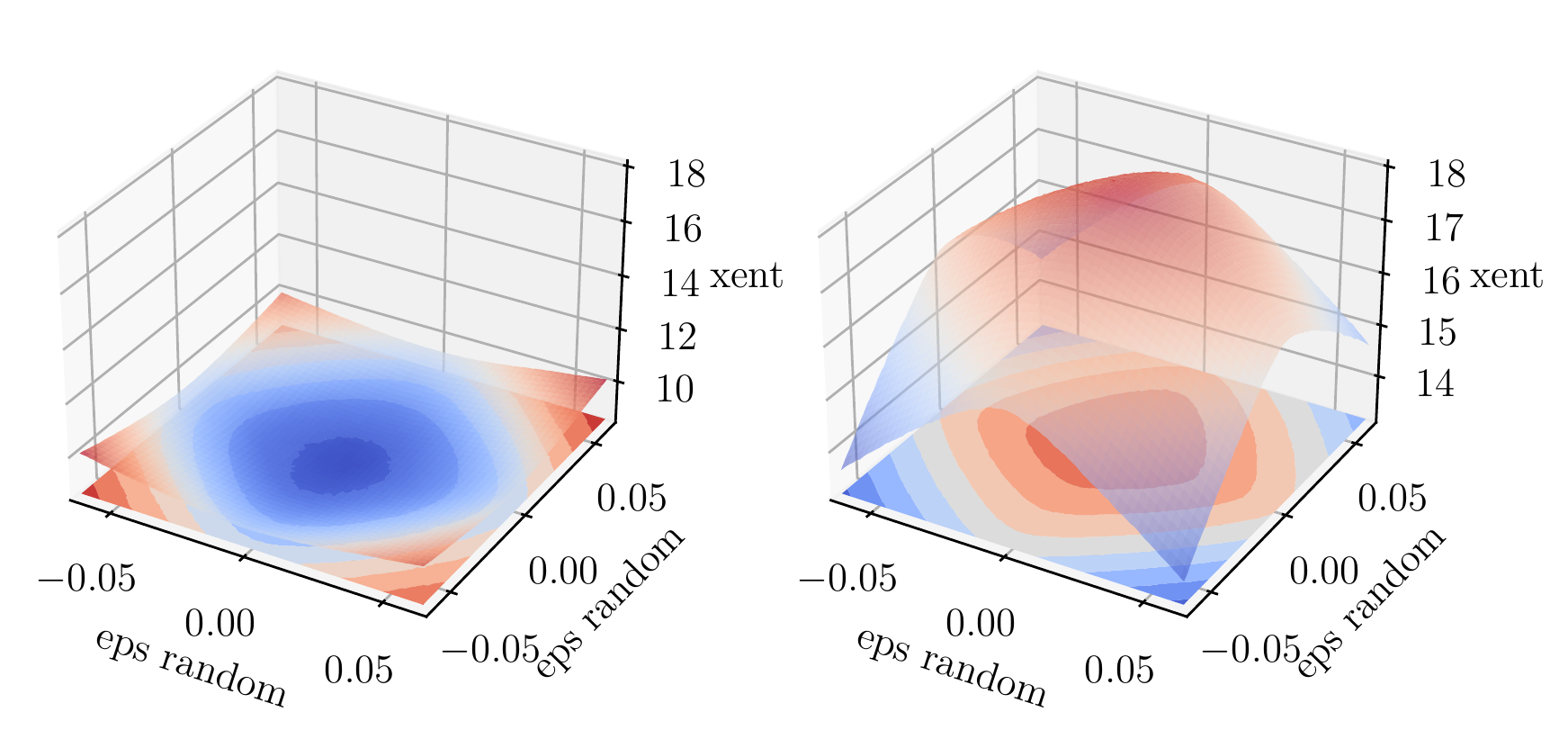}
   \vspace{-0.12in}
   \caption{\textbf{Loss surface visualization} of \system (left) and standard AT (right).}
   \label{fig:surface}
\end{figure}

\vspace{0.05in}
\noindent \textbf{Loss surface visualization.} Figure~\ref{fig:surface} visualizes the loss surface of a selected sample with both  \system (left side) and standard adversarial training (right side). On the top row, we show the cross-entropy loss projected on one random (Rademacher) and one adversarial direction. On the bottom, we project the loss along two random directions. We see that the loss surface for \system is more flat than standard adversarial training. Since both models are adversarially trained, the loss does not increase along the gradient direction. See more examples in the supplemental material.

\section{Conclusion}
We presented \system, a two-stream contrastive learning framework for improved robustness and clean accuracy. 
\system is trained using two loss functions; one to align the feature distributions between natural and adversarial images, and one to promote good classification accuracy.  The resulting model is able to defend against strong adversarial attacks at test time not only using the hardened classifier but also using a KNN search. Through extensive experiments, we demonstrate \system achieves better results than alternative methods for ImageNet under a wide range of settings.

{\small
\bibliographystyle{ieee_fullname}
\bibliography{reference}
}

\end{document}